INVESTIGATING CROSS-LINGUISTIC GENDER BIAS IN HINDI - ENGLISH ACROSS DOMAINS

SOMYA KHOSLA

LJMU MASTERS IN ML AND AI, APRIL2021

Mid-Thesis Report

OCTOBER 2021

# ABSTRACT

Measuring, evaluating and reducing Gender Bias has come to the forefront with newer and improved language embeddings being released every few months. But could this bias vary from domain to domain? We see a lot of work to study these biases in various embedding models but limited work has been done to debias Indic languages. We aim to measure and study this bias in Hindi language, which is a higher-order language (gendered) with reference to English, a lower-order language. To achieve this, we study the variations across domains to quantify if domain embeddings allow us some insight into Gender bias for this pair of Hindi-English model. We will generate embeddings in four different corpora and compare results by implementing different metrics like with pre-trained State of the Art Indic-English translation model, which has performed better at many NLP tasks than existing models.

# TABLE OF CONTENTS

# CHAPTER 1

# INTRODUCTION

With new methods and language models being generated and deployed, more and more studies are coming up to look into the social biases that have existed, are existing and how to reduce them. Social biases can be clubbed into various other sections like Religion, Ethnicity, Age, Gender, Community, etc. One of these is the Gender Bias.

Biases can derive out of previously set beliefs, preconceived notions, stereotypes, superstitions. But they do derive out of a group of people with similar thinking. This thinking can change with the times. While the mentality of a person may change, but that of a society is difficult to erase. As part of the society, the idea of bias has been inherent to us.

With this thesis, we aim to tackle one of these today.

## 1.1 Background of the Study

Gender bias has become a major focus in the past few years and recently we have seen increased studies to evaluate, generate and improve embeddings for various languages. Bias is a social construct which can be experienced by any person, in any walk of life.

Several studies have tried to study domain embeddings for various language models for measuring Gender Bias, but none, so far, have been conducted for Hindi-English language model. Through this thesis, we aim to conduct this research and investigate this bias by using domain adaptation and comparing these domain embeddings for drawing Bias results across various segments that define our social construct.

## 1.2 Problem Statement

Inferring biases in Hindi language has been an upcoming area of research. We are going to separate out our research based on two unique factors:

- We will usea relatively new architectural transformer model which is trained for Hindi-English language pairs that hasn't been used for bias studies on Hi-En before.
- We will compare these biases over domains, studying the and comparing the varying implicit biases in them.

This thesis will conduct the following:
- We will make use of IndicTrans transformer model (Ramesh et al., 2021), which will be finetuned for each domain. This will be used on domain monolingual corpora to generate domain embeddings for Hindi-English language pair.
- These embeddings will be separately evaluated for Gender bias using WEAT (Caliskan et al., 2017), TGBI metric.
- The Biases will be compared across domains to generate insights.

**1.3  Aim and Objectives**

The main aim of this research is to investigate Gender Bias across Domains in Hindi, an Indic and higher-order language (highly gendered) and compare these findings to generate insights. The research objectives are formulated based on the aim of this study which are as follows:
- To generate Hindi-English domain embeddings to draw inference on gender bias using finetuned IndicTrans model
- To generate these said embeddings for four different domains and draw inference on gender bias per domain using metrics like WEAT, TGBI, etc.
- To compare between the biases derived among the domains

**1.4  Scope of the Study**

The scope of this study is limited to:
- Generating and comparing Domain word embeddings using various metrics like WEAT, TGBI, etc.
- Generating insights on gender bias for various domains in Hindi word embeddings
- The study does not cover debiasing the word embeddings for Hindi language or for any domains.

**1.5  Significance of the Study**
- This study will provide us with insights as to how much of Gender Bias is prevalent across the Hindi language.
- This will allow us to measure how to improve the word embeddings for the language.
- Along with this, it will shed light to domain biases and how embeddings can derive value from this to reduce domain biases as well.

## 1.6 Structure of the Study

This interim thesis is divided into three chapters. Chapter 1 deals with giving us an introduction about the research problem. After a brief introduction it delves into the background of the topic giving idea about past research in the same topic and how the gender bias is perceived in Section 1.1. Section 1.2 and 1.3 introduce us to the problem statement and our objectives which will be carried out as part of our study in this research. Limitations of this research are highlighted in section 1.4 under Scope of the study. What we aim to provide through our research is mentioned briefly in Section 1.5.

We carry out the Literature Review in Chapter 2. Section 2.1 gives us a background on gender bias in Machine Translation and highlights few papers that researched this. Then we dive into Section 2.2 which talks about previous surveys and case studies to give an in-depth explanation of the domain and the problem. Under section 2.3, we discuss the most recent advancements through research in this domain. We try to raise questions from publications in domains, language translation, language models and detection as well as motivation techniques under this section in sub-sections 2.3.1, 2.3.2, 2.3.3 and 2.3.4 respectively. From our study of these papers, we try drawing comparison among various techniques used in section 2.4 and end this section after a brief summary in section 2.5.

Chapter 3 describes the methodology of this research. Section 3.1 gives us an introduction on the theme, model and metrics we will use for evaluation. Section 3.2 delves deep into it with 3.1.1 giving details about the datasets, fields and volume of data used. The next section briefly mentions the pre-processing steps for data before putting them to use in the model. Sub-section 3.2.3 gives us an architectural overview of the language model being utilised for generating embeddings. The next two sub-sections describe the methods and metrics we shall use for realizing each of our aims defined and how to evaluate for fairness using these metrics. We mention the toolkit required for carrying out our research in the next section, section 3.3 and summarize our steps in section 3.4.

# CHAPTER 2

# LITERATURE REVIEW

For a thorough review of work done in this domain, we will start with some surveys conducted and delve into the results. We will then look into related recent studies and publications that talk about gender bias in machine translation in different domains. Post that we will end the section by discussing the various techniques used to solve the problem.

## 2.1 Introduction

Many studies have been conducted for evaluating Gender Bias across various languages in Machine Translation using various methods and models. The study of Gender Bias is an old one. Starting with understanding biases in Motherhood (Güngör et al., 2009), gender stereotypes in workplace (Heilman et al., 2012), we have come to finding bias evidence in workplace, sports, media, and almost every other walk of life. Nasrina Siddiqi talks about the challenges faced by women (Saddiqi et al., 2021) over social networks, how they suffer through stereotypes on a daily basis.

Bolukbasi et al. (2016) conducted bias research in textual data using principal component analysis. Kurita et al. (2019) performed experiments on evaluating the bias in BERT using the Word Embedding Association Test (WEAT) as a baseline, which involved calculating the mean of the log probability bias score for each attribute.

More recently, we see gender bias being evaluated for Hindi-English language pair by Gupta et al. (Evaluating Gender Bias in Hindi-English Machine Translation, 2021).

We see little to no work on comparing domain gender biases.

## 2.2 Gender Bias: Surveys and Case Studies

To get a better understanding of the work done in the domain of debiasing gender roles or resolving the problem of gender bias we went through a few surveys and/or case studies. This helped us gain knowledge on what problem is in focus in recent years.

Research by Friedman et al. (Relating Linguistic Gender Bias, Gender Values, and Gender Gaps: An International Analysis) is a comprehensive analysis on gender gaps prevailing in the society. Consolidating data from multiple sources like World Values Survey data, all data was captured in English language, along with the tweets across 99 countries. The study captures multiple statistics in domains of politics, economy, education against themes or indices of gaps. This also derived the correlations of linguistics gender bias with gender valuations.

While the term bias can be led to mean anything- racial, religious, gender, etc., here it usually stems from stereotyping.

A not so recent case study of evaluating gender bias in Google Translate suggests the stereotypical biases in gender roles that show the real-world perception in roles of man and woman. Translating sentences while using English as a medium for all other languages, Google Translate also concludes an imbalance while discerning adjectives with strength towards men while leaning weaker adjectives towards women. Similarly, for professions, the translation seems to add similar adjustments by adding gendered pronouns. Check Prates et al. (2019) for further details and implementation.

A survey conducted on the increasing types of transformer-models used in NLP tasks studies the pre-trained models used and talks about various techniques and transformer architecture employed for bias mitigation. See Kalyan et al. (2021) for more.

Another survey for 'bias' conducted by Blodgett et al. (Language (technology) is power: ´ A critical survey of "bias" in NLP) conducts a survey of all publications done on studying bias in natural learning processing systems and classifies these papers according to the tasks these studies took as motivation. This went on to show that the motivation and the work done for gender bias in NLP related tasks is more than often, ambiguous and/or inconsistent.

This leads us to draw a few observations/conclusions:
- Firstly, the bias in MT systems is percolating through the inherent bias in training data or the real-world examples that we use
- While the number of studies on linguistic gender bias detection and mitigation are increasing, the term itself remains unclear as a part of motivation in the studies, i.e., the study and the motivation seem disconnected or remain unmatched.
- Most of the linguistic biases are being studied only in English. Due to this, the embeddings created in MT models from one language to another, create varied gendered outputs.

## 2.3 Recent Studies and Publications

Several language models with improved transformer architectures are now coming up that are outperforming on benchmark datasets for NLP tasks. Some of these have also offered better translation leading to improved embeddings for mitigating bias. But after having gone through the above surveys, we can pose few questions:

- Is the gender bias domain specific? Does the stereotype change when changing domain? Here domains can refer to a profession, or an industry or education, etc. So, could it be that while an overall survey for a nation co-relates woman with 'home-maker', the hospitality or wellness industry co-relates woman with adjectives like 'strong', 'driven', etc?
- What domains can we think are most affected with this bias in gender roles?
- Can language models translating to highly gendered languages create more bias if embeddings are left unchecked?
- Is debiasing embeddings the only choice we have for bias mitigation?
- Is calculating or evaluating domain adaptive biases more useful to understand?

Let us see if we can explore the answers to few of these questions ahead.

### 2.3.1 Gender Bias in Domains

Dacon and Liu (2021) ask the question- 'Does Gender matter in News'? In their study across different data in the domain of News (articles), comparative study showed immense disparity in representation. In addition to this, experiments also captured bias on basis of power, influence, career and family. As expected, adjectives connecting to men spoke of influence and decisions. While the term 'female' chimed with family, home and wedding.

Coming to journalism in sports, Fu et al. (2016) experimented with a game language model to quantify bias amongst the genders. Using the data for athletes in Tennis, they concluded that for higher ranked athletes, more game related questions were asked to male athletes when compared to their female counterparts. This study took into account questions posed in interviews to 167 male and 143 female players, showing a stark difference in the bias observed at press conference questions.

Another paper scoping women in sports and journalism observed existing marginalization of females in the domain, in few countries. This study conducted a survey of journalists for first-hand information from over 700 newspapers in countries like Australia, UK and US. While the study concludes acknowledgment of less coverage for women sports from both genders, there is still a long way to go.

These studies help us to draw a few observations:

- The percentage of female journalists in the sample was just over 5%.
- Higher number of articles about men in sports as compared to women
- In the overall dataset of articles covering both genders, females are talked about more in a domestic role.

Several studies are now also focusing on the question that whether the measurement of fairness can be adapted to domains. And if so, can it shed light on more mis-presented or under-represented categories, adjectives, stereotypes that are more connected to a particular domain?

A similar study on Wikipedia corpora was done using WEAT metric for evaluating bias across domain embeddings. The text used here belonged to English language. Using cluster embeddings, categories for bias were observed. See Chaloner and Maldonado (2019) for more.

Further research by Saunders and Byrne (2020) works on mitigating this gender bias adapting it as a domain MT problem. Using a transformer model, they compare for bias in eight diverse languages, using the base as English.

One of the most common scenarios where we see the gender bias in action is Entertainment. This is one of the videos (Johnson, A. 2021) where actress Scarlett Johansson complains about the kind of questions she is asked compared to her male counterparts while filming for movies like Marvel's Avengers. This interview depicts that even while playing roles of strong women, the more "interesting" questions flow towards the male actors.

### 2.3.2 Gender Bias in Language Translation

Now that we have understood the domain research, we shall look into the gender fairness issues in language translation. But let us put forth why this needs to be addressed for language translation:

- As already observed above, we can see that most of the language research has been carried out in English. How to measure the bias in highly gendered languages like Hindi, Spanish, etc.?

- What about languages spoken by less percentage of people, or older languages that may need to be preserved, like Sanskrit?
- Once detected, can the bias be controlled for these languages with appropriate models?

A survey conducted for Arabic language points out the same challenge of machine translation. As pointed out by Darwish et al. (2021), the dialectal data in Arabic creates challenges for many NLP applications. Thus, many such applications utilize rule-based systems.

Some work for detecting gender bias in Hindi language has been done for various domains. Kapoor, Bhuptani and Agneswaran (2017) used the Bechdel test for Hindi movies to find gendered content and biases. Madaan et al. (2018) also studies similar biases and stereotypes in Hindi movies and went one step further to mitigate these by constructing knowledge graphs. Pujari et al. (2019) attempted to debias the Hindi language using an SVM classifier.
One of the more recent papers for de-biasing gender words in Hindi by Gupta, Ramesh and Singh (2021) talk about the gendered nature of the language. This can cause embeddings to detect biased translated words while using a non-gendered or a low-order language like English.

### 2.3.3 Language Models
We'll now talk about the various model architectures used for constructing language models for analyzing and removing biases, and more importantly gender bias from the language in the recent years.

Sutskevar et al. (2014) used LSTM on the WMT'14 dataset for English to French. Using a sequence-to-sequence learning model, they experimented with reversing sentences and re-learning which gave improved results over the existing models at the time even for longer sentences, given the memory limitation of the architecture.
Vaswani et al. (2017) introduced the transformer, an attention-based model. Attention enabled inputs to interact with each other over longer length with parallelization and in turn, reduced the training time.
In 2020, Dinan et al. discussed using multiple classifiers for male, female and neutral gender categories using a pretrained transformer model by Vasvani et al. (2017), ranking the classes using a bi-encoder architecture trained with cross-entropy.

### 2.3.4 Detection and Mitigation Techniques

Gonen and Webster (2020) introduced a novel approach to detect gender bias using perturbations with BERT. This was able to automatically detect gender differences when translating the sentences in case of gendered languages.

Wong (2020) suggested to alter the data by introducing false and/or substitute data. Experiments were carried out on dataset of English and Spanish languages to see how data augmentation affects the gender bias along with the BLEU score. The data augmentation was able to mitigate the gender bias to some extent.

Word embeddings remain one of the most popular methods to detect gender bias and debiasing them is one of the ways to mitigate these biases.

### 2.5 Summary

We discussed the various papers and research to understand how gender bias is prevalent in various domains across languages. We raised some questions as to the challenges that may be present for language translation and looked into few of the solutions for that. We particularly covered a little research done for Hindi language for identifying and de-mystifying gender biases and stereotypes. Then we saw how the latest studies are trying to de-bias gendered languages like Hindi.

Post this, we discussed few recent language models and bias mitigation methods and strategies. From using SVMs and classifiers, we've come to using transformer architectures and word embeddings.

We compared a few of these and saw how few metrics are more useful to the task of gender-word ratio detection.

# CHAPTER 3

# RESEARCH METHODOLOGY

The primary aim of this thesis is to conduct gender bias investigation across domains for Hindi using En-Hi language model.

## 3.1 Introduction

Hindi is an Indic, highly gendered language with relatively less work on measuring and debiasing it. Other Indic languages are Bengali, Marathi, Punjabi, Kannada, Tamil, Assamese, Telugu, along with few more. These are the languages common the subcontinent of India. Through this, we aim to compare the biases in various domains to generate insights.

- We are to use a new SOTA transformer model for the purpose of language translation for En-Hi translation and embedding generation. This will be done separately for all four domains - sports, news, social media, entertainment
- The embeddings will then be compared based on metrics like WEAT, TGBI.

A closer analysis on the data and metrics will follow in the upcoming sections in 3.2.

## 3.2 Methodology

This section will cover details on the datasets and EDA steps. We will also discuss the machine translation language model for English to Hindi translation along with its architecture.

We will introduce two of the metrics for embedding comparison, namely WEAT and TGBI and see how it can help us in measuring gender bias in the Hindi language.

### 3.2.1 Data Selection

As mentioned, this thesis will cover data from four domains, namely- New, Sports, Social Media and Entertainment. All these datasets are publicly available on Kaggle.

The details of each dataset are as follows:

- Indian News Articles: News
- Tokyo Olympics: Sports
- Reddit Comments: Social Media

- IMDB Reviews: Entertainment

The Indian New Articles dataset contains 20k news headlines, descriptions & articles from August 11, 2019 to June 8, 2020 obtained from Indian Express. There are nine columns present We will be using 'desc' column that contains the description of the article as the source and let go of the rest of the data.

The Tokyo Olympics dataset contains over 60k records, with a total of 16 columns giving details like time, user_name, tweet_id, etc. We will be using 'user_description' column as the source for this dataset which contains the body of the tweet.

The Reddit Comments dataset has a total of 1M comments with score and controversiality for each of the 40 most frequented subreddits. We shall be using 'body' column as the source as it is the one containing body of the comment.

The IMDB Reviews dataset has 2 files, divided into test and train, with a total of 50k records. The data has just two columns: 'text', 'sentiment'. We are more concerned with the text here so we will use only the 'text' column.

### 3.2.2 Data Pre-processing

The final data we have is already clean. We'll be using the column using textual data for each of the dataset, using Pandas. We will then split this data into 80:20 ratio of train:test.

For each domain, we will finetune the language model on the training dataset and generate embeddings for Hindi language.

### 3.2.3 MT Transformer Model

IndicTrans is a Transformer-4X model trained on Samanantar dataset. Two models are available which can translate from Indic to English and English to Indic. The model can perform translations for 11 languages: Assamese, Bengali, Gujarati, Hindi, Kannada, Malayalam, Marathi, Oriya, Punjabi, Tamil, Telugu.

It offers multiple models for different language pairs: Indic-English, English-Indic and Indic-Indic. We will be using the English-Indic model to generate word embeddings.

### 3.2.4 Metrics

We shall be using two different metrics here to evaluate the bias in generated embeddings from our MT transformer model.

### 3.2.4.1 WEAT

The Word Embedding Association Test (WEAT) by Caliskan et al. (2017) is inspired by the IAT (Implicit Association Test) for word embedding models. Since machine learning algorithms like neural networks learn certain abstract features during training, the social bias of the trainings data is not only inherited but often even amplified by word embedding models (Barocas and Selbst, 2016; Zhao et al., 2017; Hendricks et al., 2018).

WEAT takes advantage of this property of word embeddings combined with the fact that the more frequently words occur together, the closer are also their vectors in the model. What WEAT does is, it takes two sets of target words of equal size and two sets of attribute embeddings. The similarity between the target sets and the attribute sets is calculated by the calculating the average cosine similarity of the words. The cosine similarity captures the cosine of the angle between two vectors ~x and ~y which works for any number of dimensions.

We define the similarity between the two vectors by their Cosine function. The more similar they are, the below value will attain a maximum value of 1 for 0°. This similarity metric is defined as follows:

$$\text{sim}_{\text{Cosine}}(\vec{x}, \vec{y}) = \frac{\vec{x} \cdot \vec{y}}{\|\vec{x}\| \cdot \|\vec{y}\|} = \frac{\sum_{j=1}^{m} \vec{x}_j \cdot \vec{y}_j}{\sqrt{\sum_{j=1}^{m} \vec{x}_j^{\,2}} \cdot \sqrt{\sum_{j=1}^{m} \vec{y}_j^{\,2}}}$$

To calculate the significance, p-value is calculated via performing permutation test. It is important to note here that though WEAT was inspired by IAT, the p-value here is different than reported originally in IAT.

### 3.2.4.1 TGBI

The Translation Gender Bias Index (TGBI) is a measure to detect and evaluate the gender bias in MT systems, introduced by Cho et al. (2019). They use Korean-English (KN-EN) translation. In Cho et al. (2019), the authors create a test set of words or phrases that are gender neutral in the source language, Korean. These lists were then translated using three different models and evaluated for bias using their evaluation scheme. The evaluation methodology proposed in the paper quantifies associations of 'he,' 'she,' and related gendered words present in translated text.

### 3.2.5 Evaluation

Let us now delve into how we will realize each of our aims.

Using the English-Indic model of IndicTrans, we will generate domain embeddings for English and Hindi on the training data.

Once we have the embeddings, we will compare these embeddings on the test data per domain, to derive the gender gaps in the language.

Once we have these statistics for all four domains, we will be able to draw comparison on both genders and see if we can answer our previous questions.

We shall ponder over these:

Do the embeddings for male: female vary for each domain?

Where can we see dominance of one gender over the other?

For each domain, which are the words that are more closely related to a gender?

## 3.3 Tools

For completing the research project, we would require:

- Platform

A dedicated DL platform GPU enabled for training transformer on huge corpuses for next 3 months as free Google Colab won't allow to train this huge model (>400M params) like AWS

- Other libraries used:

Python

Pandas

Numpy

## 3.4 Summary

This section started with giving an introduction on the languages and aims in section 3.1. Under section 3.2, we laid out our datasets, their high-level description and the data which we will use going forward. The next two subsections delved into the data pre-processing steps and providing architectural understanding over the transformer model we are using. Section 3.2.4 describes the metrics we are using for drawing comparison over the domain embeddings in conjunction with next subsection that will tell us how to evaluate using these metrics.

In section 3.3, we are listing the tools we'll require to complete our research, ending with summarizing our methodology in section 3.4.

**References**

Caliskan, Aylin, Joanna J. Bryson, and Arvind Narayanan (2017). „Semantics derived automatically from language corpora contain human-like biases". In: Science 356.6334, pp. 183–186.

Madeline E. Heilman (2012), Gender stereotypes and workplace bias. In: Research in Organizational Behavior 32, pp.113-135

Güngör, G., Biernat, M. Gender Bias or Motherhood Disadvantage? Judgments of Blue-Collar Mothers and Fathers in the Workplace. Sex Roles 60, 232–246 (2009)

Siddiqi, N., 2021. Self-Expression in the Cyber World: Challenges for a Woman. Indian Journal of Gender Studies, p.09715215211030586.

Ramesh, G., Doddapaneni, S., Bheemaraj, A., Jobanputra, M., AK, R., Sharma, A., Sahoo, S., Diddee, H., Kakwani, D., Kumar, N. and Pradeep, A., 2021. Samanantar: The Largest Publicly Available Parallel Corpora Collection for 11 Indic Languages. arXiv preprint arXiv:2104.05596.

Bolukbasi, T., Chang, K.W., Zou, J.Y., Saligrama, V. and Kalai, A.T., 2016. Man is to computer programmer as woman is to homemaker? debiasing word embeddings. Advances in neural information processing systems, 29, pp.4349-4357.

Kurita, K., Vyas, N., Pareek, A., Black, A.W. and Tsvetkov, Y., 2019. Measuring bias in contextualized word representations. In Proceedings of the First Workshop on Gender Bias in Natural Language Processing, Association for Computational Linguistics, pp 166–172

Won Ik Cho, Ji Won Kim, Seok Min Kim, and Nam Soo Kim. 2019. On measuring gender bias in translation of gender-neutral pronouns. In Proceedings of the First Workshop on Gender Bias in Natural Language Processing, pages 173–181, Florence, Italy. Association for Computational Linguistics

Barocas, S. and Selbst, A.D., 2016. Big data's disparate impact. Calif. L. Rev., 104, p.671.

Zhao, J., Wang, T., Yatskar, M., Ordonez, V. and Chang, K.W., 2017. Men also like shopping: Reducing gender bias amplification using corpus-level constraints. In Proceedings of the 2017 Conference on Empirical Methods in Natural Language Processing.

Hendricks, L.A., Burns, K., Saenko, K., Darrell, T. and Rohrbach, A., 2018. Women also snowboard: Overcoming bias in captioning models. In Proceedings of the European Conference on Computer Vision (ECCV) (pp. 771-787).

Darwish, K., Habash, N., Abbas, M., Al-Khalifa, H., Al-Natsheh, H.T., El-Beltagy, S.R., Bouamor, H., Bouzoubaa, K., Cavalli-Sforza, V., El-Hajj, W. and Jarrar, M., 2021. A Panoramic Survey of Natural Language Processing in the Arab World. Communications of the ACM v.64, (4) pp 72-81

Kapoor, H., Bhuptani, P.H. and Agneswaran, A., 2017. The Bechdel in India: gendered depictions in contemporary Hindi cinema. Journal of Gender Studies, 26(2), pp.212-226.

Madaan, N., Mehta, S., Agrawaal, T., Malhotra, V., Aggarwal, A., Gupta, Y. and Saxena, M., 2018, January. Analyze, detect and remove gender stereotyping from bollywood movies. In Conference on fairness, accountability and transparency (pp. 92-105). PMLR.

Gupta, G., Ramesh, K. and Singh, S., 2021. Evaluating Gender Bias in Hindi-English Machine Translation. arXiv preprint arXiv:2106.08680.

Blodgett, S.L., Barocas, S., Daumé III, H. and Wallach, H., 2020. Language (technology) is power: A critical survey of" bias" in nlp. arXiv preprint arXiv:2005.14050.

Friedman, S., Schmer-Galunder, S., Chen, A., Goldman, R. and Rye, J., 2019. Relating linguistic gender bias, gender values, and gender gaps: An international analysis. In BRiMS Conference, Washington, DC, July.

Dacon, J. and Liu, H., 2021, April. Does Gender Matter in the News? Detecting and Examining Gender Bias in News Articles. In Companion Proceedings of the Web Conference 2021 (pp. 385-392).

Liye, F., Danescu, C. and Lee, L., 2016. Tie-breaker: Using language models to quantify gender bias in sports journalism. Computation and Language.

Schmidt, H.C., 2018. Forgotten athletes and token reporters: Analyzing the gender bias in sports journalism. Atlantic Journal of Communication, 26(1), pp.59-74.

Kalyan, K.S., Rajasekharan, A. and Sangeetha, S., 2021. Ammus: A survey of transformer-based pretrained models in natural language processing. arXiv preprint arXiv:2108.05542.

Prates, M.O., Avelar, P.H. and Lamb, L.C., 2020. Assessing gender bias in machine translation: a case study with google translate. Neural Computing and Applications, 32(10), pp.6363-6381.

Saunders, D. and Byrne, B., 2020. Reducing gender bias in neural machine translation as a domain adaptation problem. arXiv preprint arXiv:2004.04498.

Dinan, E., Fan, A., Wu, L., Weston, J., Kiela, D. and Williams, A., 2020. Multi-dimensional gender bias classification. arXiv preprint arXiv:2005.00614.

Gonen, H. and Webster, K., 2020. Automatically identifying gender issues in machine translation using perturbations. arXiv preprint arXiv:2004.14065.

Wong, A., 2020. Mitigating Gender Bias in Neural Machine Translation Using Counterfactual Data.

Johnson, A., 2020. scarlett johansson shutting down sexist comments for 5 min straight [online video]
Available at: https://www.youtube.com/watch?v=YGqQk12jBoA